%% file: main.tex
\documentclass[runningheads]{llncs}
\usepackage[T1]{fontenc}
\usepackage{graphicx}
\usepackage{subcaption}
\usepackage{pdflscape}
\usepackage{subcaption}
\usepackage{tabularx}
\usepackage{booktabs} 
\usepackage{siunitx}  
\usepackage{tabularx}
\usepackage{makecell}
\usepackage{booktabs}
\usepackage{siunitx}
\begin{document}
\title{Evaluating YOLO Architectures: Implications for Real-Time Vehicle Detection in Urban Environments of Bangladesh}
\titlerunning{Evaluating YOLO Architectures}
%
\author{Ha Meem Hossain\inst{1}\orcidID{0009-0005-8188-1616} \and
Pritam Nath\inst{1}\orcidID{0009-0005-4830-4614} \and
Mahitun Nesa Mahi\inst{1}\orcidID{0009-0001-1096-1169} \and
Imtiaz Uddin\inst{1}\orcidID{0009-0009-5748-2104} \and
Ishrat Jahan Eiste\inst{1}\orcidID{0009-0009-7141-1935} \and
Syed Nasibur Rahman Ratul\inst{1}\orcidID{0009-0002-7468-7098} \and
Md Naim Uddin Mozumdar\inst{1}\orcidID{0009-0000-9705-837X} \and
Asif Mohammed Saad\inst{1}\orcidID{0009-0006-1315-6444}\thanks{Corresponding author: Asif Mohammed Saad (orcid: 0009-0006-1315-6444)}
\and
MD Tamim Hossain\inst{1}\orcidID{0009-0007-9421-3494}
}
\authorrunning{H. Hossain et al.}

%
\institute{Department of Computer Science and Engineering,\\
Premier University, Chittagong, 1/A O.R. Nizam Rd, Chattogram 4203, Bangladesh
\email{hameemhossain2@gmail.com, pritamnath1971@gmail.com, mahitunnesamahi11@gmail.com, mdimtiaz1048@gmail.com, eisteishrat@gmail.com, sayednasiburrahman@gmail.com, mdnaimuddinmozumdar@gmail.com,
asifsaad730@gmail.com,
thossain33333@gmail.com}
}

\maketitle              
\begin{abstract}
Vehicle detection systems trained on Non-Bangladeshi datasets struggle to accurately identify local vehicle types in Bangladesh's unique road environments, creating critical gaps in autonomous driving technology for developing regions. This study evaluates six YOLO model variants on a custom dataset featuring 29 distinct vehicle classes, including region-specific vehicles such as ``Desi Nosimon'', ``Leguna'', ``Battery Rickshaw'', and ``CNG''. The dataset comprises high-resolution images (1920×1080) captured across various Bangladeshi roads using mobile phone cameras and manually annotated using LabelImg with YOLO format bounding boxes. Performance evaluation revealed YOLOv11x as the top performer, achieving 63.7\% mAP@0.5, 43.8\% mAP@0.5:0.95, 61.4\% recall, and 61.6\% F1-score, though requiring 45.8 milliseconds per image for inference. Medium variants (YOLOv8m, YOLOv11m) struck an optimal balance, delivering robust detection performance with mAP@0.5 values of 62.5\% and 61.8\% respectively, while maintaining moderate inference times around 14-15 milliseconds. The study identified significant detection challenges for rare vehicle classes, with Construction Vehicles and Desi Nosimons showing near-zero accuracy due to dataset imbalances and insufficient training samples. Confusion matrices revealed frequent misclassifications between visually similar vehicles, particularly Mini Trucks versus Mini Covered Vans. This research provides a foundation for developing robust object detection systems specifically adapted to Bangladesh traffic conditions, addressing critical needs in autonomous vehicle technology advancement for developing regions where conventional generic-trained models fail to perform adequately.

\end{abstract}
\section{Introduction}
\input{sections/introduction}

\section{Related Works}
\input{sections/related_works}

\section{Methodology}

\input{sections/methodology}

\section{Result}
\input{sections/result}

\section{Conclusion}
\input{sections/conclusion}

\begin{credits}

\end{credits}
%
%
%
%





\end{document}

%% file: sections/introduction.tex
 The rapid evolution of artificial intelligence and computer vision has unlocked transformative possibilities across numerous sectors, with object detection emerging as one of the most impactful \cite{Reference_12}. Object detection helps classify and identify various objects to automate various tasks via Artificial Intelligence \cite{Reference_14,Reference_20}. Various systems are being made for different industries, such as Corp monitoring for Agriculture \cite{Reference_8}, Crowd monitoring for Security and Surveillance, Trash Detection \cite{Reference_21}, Patient Monitoring for Healthcare \cite{Reference_13,Reference_23,Reference_22}
, Autonomous Vehicle System for Automobile, etc.

 A cornerstone of developing such systems is accurate vehicle and object detection on the road and identification. Many Computer vision systems related to traffic or automobiles need an accurate idea of the road in real time. Although sufficient research and training have been conducted to develop models with sufficient data to accurately identify the road scenes of that part of the world, other parts of the world have remained relatively overshadowed by such systems \cite{Reference_3,Reference_10}. Our goal is to help fill the gap by training models with images taken from a specific region of Bangladesh and comparing them to determine which can best detect the road scene of Bangladesh with its unique environment of the road. Also, Real-time object detection for autonomous vehicles and other unmanned systems significantly enhances road safety and contributes to the advancement of the automotive industry \cite{Reference_19,Reference_7}. 
 
However, achieving precise and reliable vehicle and object detection remains a complex task, facing numerous challenges. These include variations in lighting conditions, dense occlusions, the wide diversity of vehicle types and sizes, different viewing angles, and the quality of images or video feeds \cite{Reference_6}. Detecting small or irregularly shaped objects presents a particular difficulty, as many current models are primarily designed to focus on larger objects \cite{Reference_8}. Furthermore, implementing object detection systems in environments with irregular traffic patterns, cluttered scenes, and unique local vehicle types, such as those found in densely populated countries like Bangladesh, adds layers of complexity and highlights the need for high-quality, representative datasets \cite{Reference_4}.

 To help solve this problem, we are in need of a large amount of data on Bangladeshi Road scenes \cite{Reference_15}. Although there are now a lot of road environment datasets from Bangladesh, ours is distinct from others. Based on the local automobiles found on Bangladeshi roads, we have separated our dataset into 29 distinct classes, with the majority of the classes being named after local vehicles. These consist of ``Desi Nosimon'', ``Leguna'', ``Battery Rickshaw'', ``Animal'', ``CNG'', and more. Photographs are manually labeled, arranged, and trained using the YOLOv8 and YOLOv11 models in three different sizes (n, m, and x). Our dataset improves the model's capacity to detect and distinguish the unique vehicles and road features found in Bangladesh.
 
The main contribution of this work is to the improvement of Bangladeshi road condition detection. There are not many available object detection datasets, or we can say those datasets are limited; however, our internal dataset offers a strong basis. Which can contribute in some scenarios, like:

- The research aims to improve traffic management by integrating real-time vehicle detection for congested traffic sections of the road by automating or helping traffic control.

- It could speed up the process for autonomous driving in harsh and unpredictable regions by training different real-time object (vehicle) models and comparing them using a native dataset model.

%% file: sections/related_works.tex
Recent research has concentrated on enhancing object detection via various models such as YOLO or R-CNN to make them faster, more accurate, and compatible with real-time applications. Here we discuss research that evaluates the performance of different models for object detection on different datasets, specifying different goals and findings, all aiming at the research for object detection or vehicle detection.

Hasib Zunair et al.\cite{Reference_3} introduce RSUD20K, a newly developed, massive dataset that trains models to identify complex road scenes in Bangladesh. The dataset includes different scenarios, 20,000 high-quality images captured from dashboard cameras, and 130,000 annotations for 13 objects. They examined the dataset and assisted with labeling by employing well-known object detection models such as YOLOv6, YOLOv8, and larger models such as Grounding DINO and OWL-ViT. YOLOv6-L performed the best in their tests, with an average precision (mAP) of 73.7\%. The results indicated that this dataset is more difficult than many others.  

To properly evaluate the dataset, Nowrin et al. \cite{Reference_1} uses the RSUD20k dataset trained with three different versions of the YOLO-NAS model (S, M, and L) to help autonomous vehicles in Bangladesh better identify objects.Out of the three, YOLO-NAS-M gave the best results with 82.6\% mAP@0.5, beating other models like Faster R-CNN and YOLOv8. The study discovered that YOLO-NAS does a good job of finding different vehicles and people in busy, complicated places.

The purpose of Martinus Grady Naftali et al. \cite{Reference_2}  is to determine which object detection models are best for recognizing cars, people, and signs in street images, which is important for self-driving cars. They used a modified Udacity Self-Driving Car dataset that contained 3,169 photos and 24,102 annotations.  They only have five subject classes.  They ran this data through different models like SSD MobileNetv2, YOLOv3, YOLOv4, YOLOv5s, and YOLOv5l.  YOLOv5l had the best accuracy (mAP@0.5 = 0.593), but MobileNetv2 was the fastest.  It was shown that YOLOv5s is a good balance of speed and accuracy.

Another study Sundaresan Geetha et al. \cite{Reference_7}looks at two versions of a well-known object detection model, YOLOv8 and the newer YOLOv10, on a public GitHub dataset of 1,321 tagged photos of vehicles to see which one is better at finding vehicles. They say that YOLOv10 was better than YOLOv8 when it came to speed and accuracy of detection. The results show that YOLOv10 is better for real-time vehicle detection tasks like traffic monitoring and surveillance systems. 

In a study of Ali et al. \cite {Reference_4} on real-time object detection for autonomous vehicles in Bangladesh, a YOLOv5 model was trained using transfer learning on a custom Roboflow dataset of 6351 images from Dhaka and Chittagong with 23 classes. Achieved on Google Colab Pro, the model obtained 95.8\% accuracy and 0.3 ms processing time per image. This demonstrated effective learning and robust performance for autonomous vehicles in urban environments.

Al Mudawi et al. \cite{Reference_5} introduce a real-time vehicle detection system designed for drones, mainly using YOLOv8. The process starts with image enchantments like de-fogging. Following this, it uses Fuzzy C-Means to separate vehicles from the background and features are extracted using a SIFT technique. This system was evaluated on two datasets VEDAI and VAID achieving accuracies of 95.6\% and 94.6\% respectively.

Telaumbanua et al. \cite{Reference_6} introduces a real-time system for detecting and tracking nine types of vehicles using YOLOv8 for detection and Deep Sort for tracking.The YOLOv8 model was trained on a dataset of CCTV footage , consisting of 2,042 training images, 612 testing images, with 100 epochs. And it achieved a 96\% accuracy on test data. This trained model is capable of detecting vehicles in both images and live video streams.

Talib et al.\cite{Reference_8} introduced YOLOv8-CAB,  tuned to spot small and low-visibility objects. What is novel here is the combination of the Context Attention Block and spatially optimized Spatial Attention Modules that improve feature extraction and allow the model to pick up weak signals in chaotic or noisy images. The model was tested o and reached an average mAP of 97\% , outperforming baseline YOLOv8 and other light models like YOLOv5 and YOLO-NAS. The approach demonstrates the significance of architectural improvements in improving YOLO models in cluttered environments. 

In another paper, Hsu et al. \cite{Reference_9}, a simplified Fast R-CNN framework for improving the efficiency of vehicle detection systems. Their approach removes duplicate classification layers to conserve computation time and training time. Through the utilization of the SHRP 2 NDS dataset, their model detected vehicles from different angles in varied conditions. The system achieved over 90\% accuracy and up to 98.5\% recall, with strong performance on real-world surveillance video streams. Though not South Asia-focused, this work underscored the importance of model simplification and fine-tuning to improve detection performance under resource-constrained conditions. 

Bipin Saha et al.\cite{Reference_10} recognized the absence of vehicle detection research for South Asian conditions and introduced the Bangladeshi Native Vehicle Dataset (BNVD). The dataset contains over 81,000 annotated samples across 17,326 images, spanning 17 native vehicle classes like rickshaws, bhotbhoti, easybikes, and vans. Collected from 17 Bangladeshi districts, the images span various weather, lighting, and orientations of vehicles to enhance model robustness. The assessment was conducted for YOLOv5, YOLOv6, YOLOv7, and YOLOv8, of which YOLOv8 did the best, with mean  mAP@0.5 being 84.8\% and mAP@[0.5:0.95] being 64.3\%.

%% file: sections/methodology.tex
The methodology section has been divided into four subsections. (i) Data Acquisition and Preparation: describes how data was acquired and prepared, (ii) Model Architecture: Model selection based on the model's architecture, (iii)Training Procedure: describes how and on which parameter the model was trained, (iv) Evaluation Metrics: how the model's performance will be evaluated.
\subsection{Data Acquisition and Preparation}
In order to accurately document real-world road conditions, we used mobile phone cameras. To ensure that every image had the same resolution and to accelerate the labeling procedure, we used the ``Open Camera'' application from the Google Play Store and set the resolution to 1920*1080 pixel. This method helps to maintain uniform image quality throughout our dataset. We took pictures of different roads in Bangladesh to document a variety of vehicles, objects, and traffic patterns. Our goal was to include all 29 classes that we had set out to categorize. For convenience and storage, all collected images were saved in a shared Google Drive folder. The complete data sets were hand-annotated to localize objects of interest with bounding boxes and detect them. Annotation was performed using LabelImg, an open-source graphical image annotation tool, and was installed and run in an Anaconda environment for better package management as well as cross-platform compatibility. Anaconda provided an expandable Python environment, which made the installation of dependencies straightforward and ensured consistent behavior across systems.
All images were annotated by drawing bounding boxes around objects and marking classes according to a prespecified list of 29 various object classes, including generic and region-specific vehicle types (e.g., CNGs, rickshaw-vans, legunas, nosimons) and non-vehicle objects like people and animal.
Annotations were saved in the YOLO format, where each object is represented by a single line in a .txt file.
\subsection{Model Architecture}
\begin{figure}[h]
    \centering
    \includegraphics[width=0.9\textwidth]{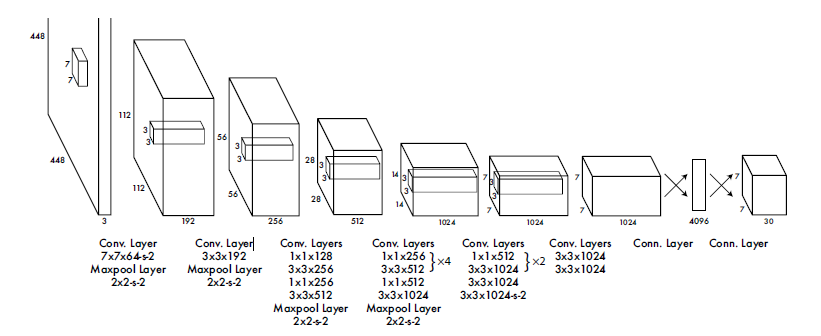}
    \caption{YOLO Model Architecture}
    \label{fig:yolo architecture}
\end{figure}

In recent years, as the need for object detection has arisen and gradually increased, new models with different architectures have emerged, such as YOLO, R-CNN, and SSD. Among these, YOLO(You Only Look Once), a family of object detector models, has stood out from other models for its speed-accuracy trade-off in real-time applications. Unlike other models, which are region-based and rely on multiple stages, YOLO does detection as a single regression problem to process an entire image in one forward pass, allowing real-time detection with low computation as presented in figure \ref{fig:yolo architecture}. 
With the context of our used dataset of road scenes mainly targeting vehicles, which has a large class, and some classes have few instances to train on. We selected YOLOv8 and YOLOv11. YOLOv8 uses C2f modules, SPPF(Spatial Pyramid Pooling-Fast), and decoupled, which makes better generalization and faster convergence \cite{Reference_16}. Thus enabling localize overlapping and small objects effectively. And YOLOv11 uses components such as SCDown modules, C3k2 blocks, and Large kernel Depthwise Convolutions, which improve semantic richness and context awareness in scenes like vehicle detection \cite{Reference_17}. We chose to experiment with ``n'',``m'', and ``x'' variants of YOLOv8 and 11. ``n'' for nano models are light and optimized for inference on low-resource devices. ``m'' for medium models offer a balanced trade-off, while ``x'' for extra large models provide maximum power, potentially making them more adept at learning from classes with fewer instances \cite{Reference_18}.

\subsection{Training Procedure}
To train the images and then validate the trained model, the entire dataset was divided into 3 sections. 75\% was allocated for training. 5\% was for validation during training after each epoch. And 20\% was allocated as a testing dataset. During training, for each model and its variants, parameters related to training models, such as epoch, workers, seed, initial learning rate, and end learning rate etc. were tweaked to get maximum output from training. Also, batch size was set to -1 to use the maximum available GPU at a single time. And the image size was set to 640.

\subsection{Evaluation Metrics}
To evaluate the comparison of the models used, values of some evaluation metrics have to be considered, such as Recall, F1 score, mAP@0.5, mAP@0.5:0.95, inference, etc.
Recall is the metric that measures the model's ability to correctly detect all relevant objects, in our case, vehicles, and it is computed as the ratio of true positives to the total number of ground truth instances. 
\begin{equation}
\text{Recall} = \frac{TP}{TP + FN}
\end{equation}
A higher recall indicates the model is less likely to miss objects which is really important for real-time vehicle detection.
F1 Score, used in the YOLO training log, is the harmonic mean of precision and recall, giving a balanced view of the model's classification performance.
\begin{equation}
\text{F1} = 2 \cdot \frac{\text{Precision} \cdot \text{Recall}}{\text{Precision} + \text{Recall}}
\end{equation}
mAP@0.5 is one of the primary metrics for object detection. It calculates the average precision across all classes using a fixed IoU threshold of 0.5. 
\begin{equation}
\text{mAP@0.5} = \frac{1}{N} \sum_{i=1}^{N} \text{AP}_i^{\text{IoU} = 0.5}
\end{equation}
mAP@0.5:0.95 is the mean of AP scores calculated at 10 IoU thresholds from 0.5 to 0.95. This metric provides a more detailed assessment of localization accuracy.
\begin{equation}
\text{mAP@0.5:0.95} = \frac{1}{10} \sum_{j=0}^{9} \left( \frac{1}{N} \sum_{i=1}^{N} \text{AP}_i^{\text{IoU} = 0.5 + 0.05 \cdot j} \right)
\end{equation}
Inference time refers to the average time(in ms) required to process a single image during inference. This is a crucial metric to distinguish which model will perform better in a real-time application.
Class-wise AP@0.5 is the average precision for each individual class at an IoU threshold of 0.5. This metric will help analyze models to determine if they can identify rare object types accurately.

%% file: sections/result.tex
This section is divided into four parts: (i) Quantitative Performance Analysis – presents and explains the overall numerical outcomes of the evaluated models, (ii) Qualitative Detection Results – provides a class-wise analysis of each model's performance on the dataset along with interpretive insights, (iii) Model Performance – evaluates and discusses the models' effectiveness and practicality based on quantitative results, (iv) Critical Assessment – offers a detailed analysis and concluding remarks on the models' performance across different classes.

\subsection{Quantitative Performance Analysis}
\begin{figure}[htbp]
    \centering
    \begin{tabular}{cc}
        \subcaptionbox{Models' Classwise Performance on AP@0.5 metrics\label{fig:classwise_ap_barchart}}
            {\includegraphics[width=0.5\linewidth]{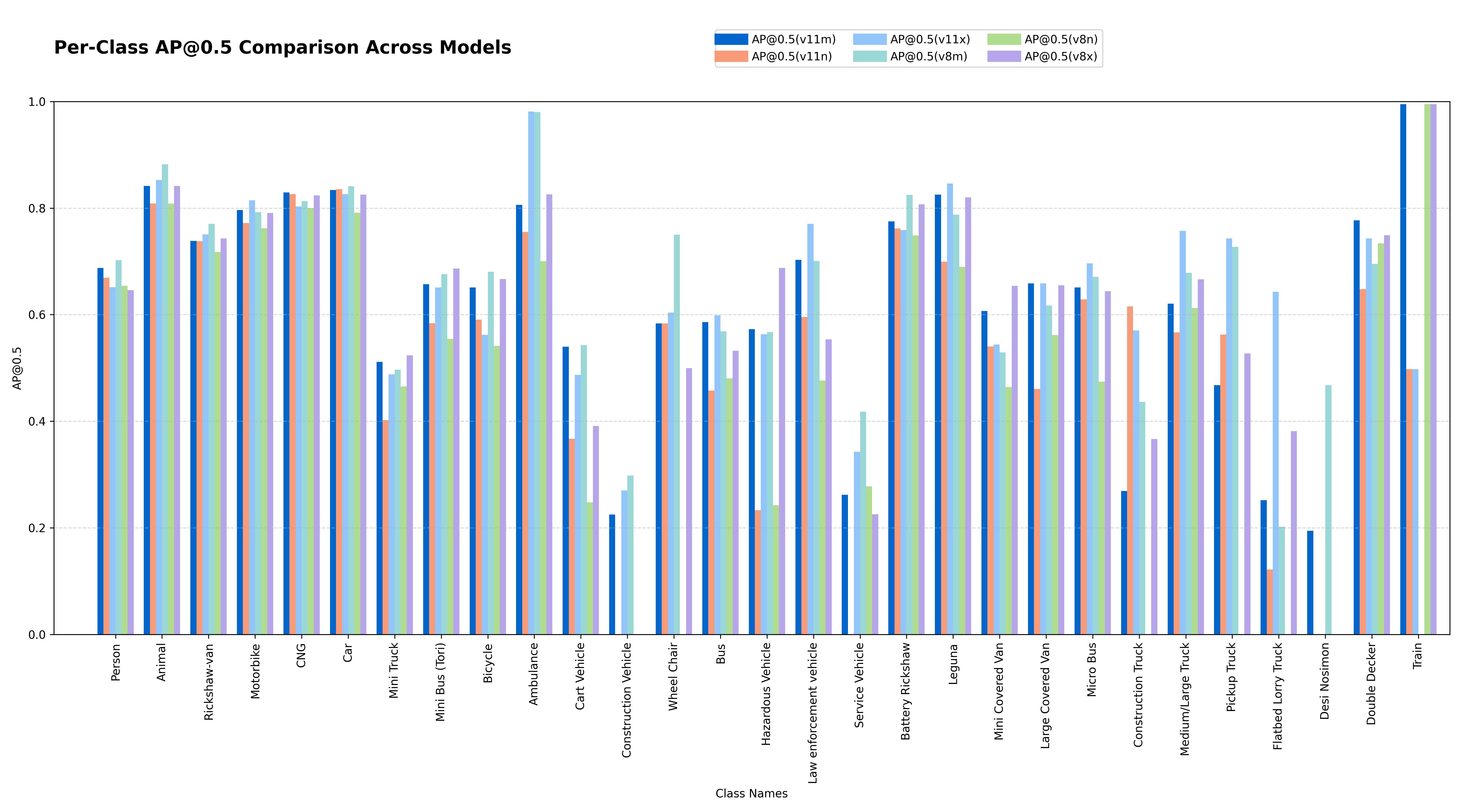}}
        &
        \subcaptionbox{Models' Performance on basis metrics\label{fig:model_metrics_chart}}
            {\includegraphics[width=0.5\linewidth]{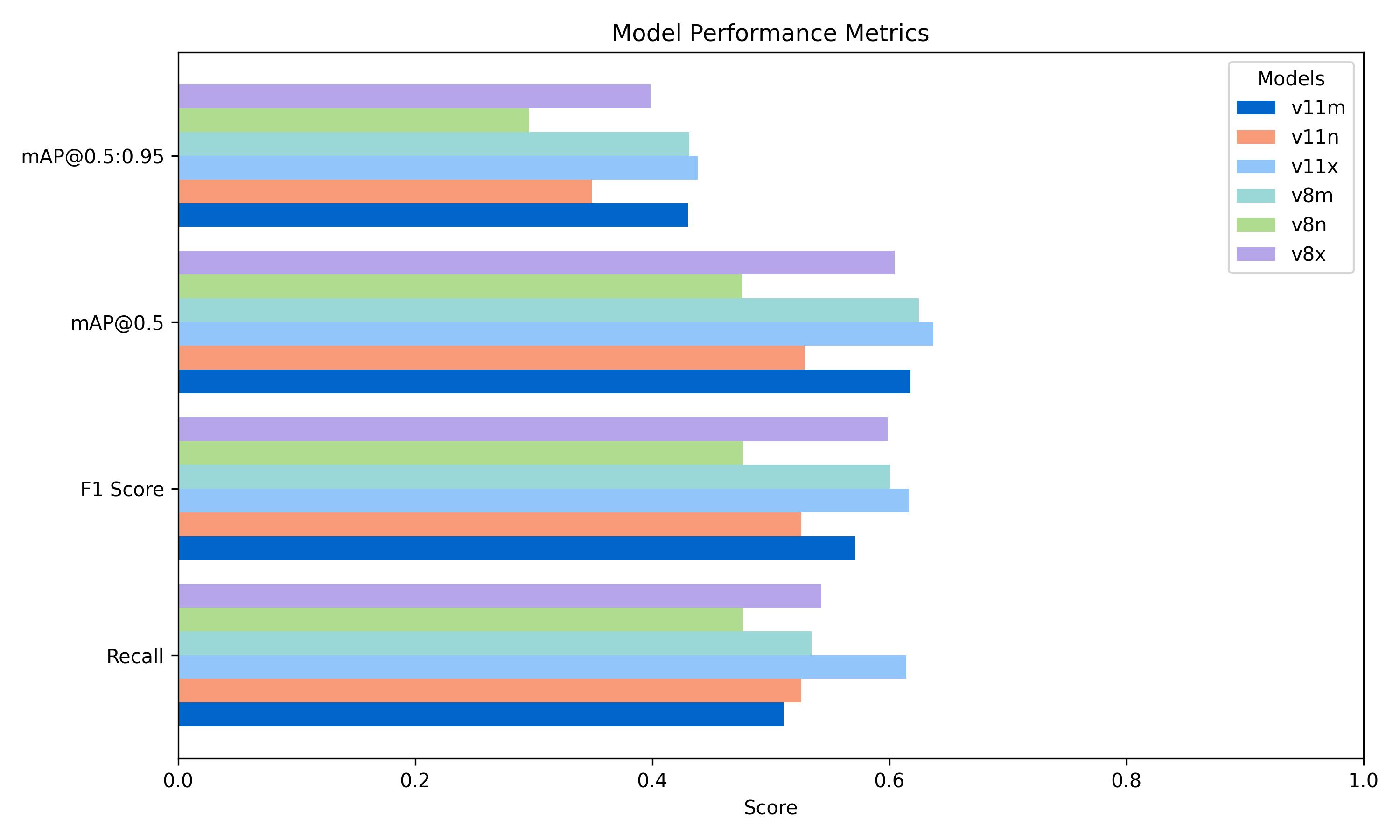}}
    \end{tabular}
    \caption{Evaluation of 6 variants of YOLO models across 4 different metrics}
    \label{fig:yolo_comparison}
\end{figure}

The performance of all three variant models of YOLOv8 and YOLOv11, which combine six models, is visualized in Figure~\ref{fig:classwise_ap_barchart}. It provides a clear snapshot of how each model performs across different object classes, using AP@0.5 as the metric. For common and visually distinctive classes like Person, Car, CNG, and Double Decker, all six models achieve strong results, often with AP scores above 0.8. This suggests that these categories are well-represented in the data and are easier for the models to detect. However, the story is quite different for challenging or less frequent classes such as Construction Vehicle, Hazardous Vehicle, Desi Nosimon, and Law Enforcement Vehicle. Here, AP scores drop noticeably—sometimes below 0.5, regardless of the model, highlighting persistent difficulties in recognizing these objects. Among the models, YOLOv11x or  v11x generally stands out with slightly higher AP values for many classes, especially for Mini Bus (Tori), Bus, and Train. Still, for the toughest categories, even the best models struggle. 

Figure~\ref{fig:model_metrics_chart} visually compares the performance of the six models across four key metrics: mAP@0.5, mAP@0.5:0.95, Recall, and F1 Score. The mAP metrics are especially important for assessing detection accuracy, as they reflect model performance over a range of Intersection over Union (IoU) thresholds from 0.5 to 0.95. Among all models, YOLOv11 XL stands out as the top performer, achieving an mAP@0.5 of approximately 0.637 (63.7\%) and an mAP@0.5:0.95 of about 0.438 (43.8\%). Additionally, this model leads in both Recall (61.4\%) and F1 Score (61.6\%), as detailed in Table~\ref{tab:model_metrics}.

\vspace{-10pt}

\begin{table}[ht]
\centering
\setlength{\tabcolsep}{3pt}
\begin{tabularx}{\linewidth}{l *{6}{S[table-format=1.3]} S[table-format=2.2]}
\toprule
Model & {IoU} & {Precision} & {Recall} & {F1 Score} & {mAP@0.5} & {\makecell{mAP\\@0.5:0.95}} & {\makecell{Inference\\Time (ms/img)}} \\
\midrule
v8n   & 0.296 & 0.476 & 0.477 & 0.476 & 0.476 & 0.296 & 2.70  \\
v8m   & 0.431 & 0.685 & 0.534 & 0.600 & 0.625 & 0.431 & 14.29 \\
v8x   & 0.399 & 0.667 & 0.543 & 0.599 & 0.604 & 0.399 & 33.55 \\
v11n  & 0.349 & 0.526 & 0.526 & 0.526 & 0.528 & 0.349 & 2.52  \\
v11m  & 0.430 & 0.647 & 0.511 & 0.571 & 0.618 & 0.430 & 14.84 \\
v11x  & 0.438 & 0.619 & 0.614 & 0.617 & 0.637 & 0.438 & 45.83 \\
\bottomrule
\end{tabularx}
\caption{Main performance metrics for all model variants.}
\label{tab:model_metrics}
\end{table}

\subsection{Qualitative Detection Results}
In order to compare the models' performance, a sample image containing 5 objects, of which 2 are motorbikes, 1 Car, 1 CNG, and 1 Mini Covered van, is selected. With Figure~\ref{fig:yolo_model} serving as the reference image for assessing performance, among the six YOLO model variants, YOLOv8n correctly detected 3 objects, of which 2 were Motorbikes and 1 CNG, but failed to detect the Car and falsely detected the Mini Covered Van as a Mini Truck, a different class. YOLOv8m failed to detect the Mini Covered Van but successfully detected the other 4 objects. YOLOv8x perfectly detected all 5 objects with a good confidence score for each object. Then, the ``n'' variant of YOLOv11 also detected all objects perfectly with all good confidence scores except the object of class CNG. YOLOv11m failed to detect two objects, but perfectly detected 2 motorbikes and 1 CNG. Lastly, the ``x'' version of YOLOv11, which is the most powerful model among all models, successfully detected all 5 objects.

\begin{figure}[htbp]
    \centering
    \begin{subfigure}{0.3\linewidth}
        \includegraphics[width=\linewidth]{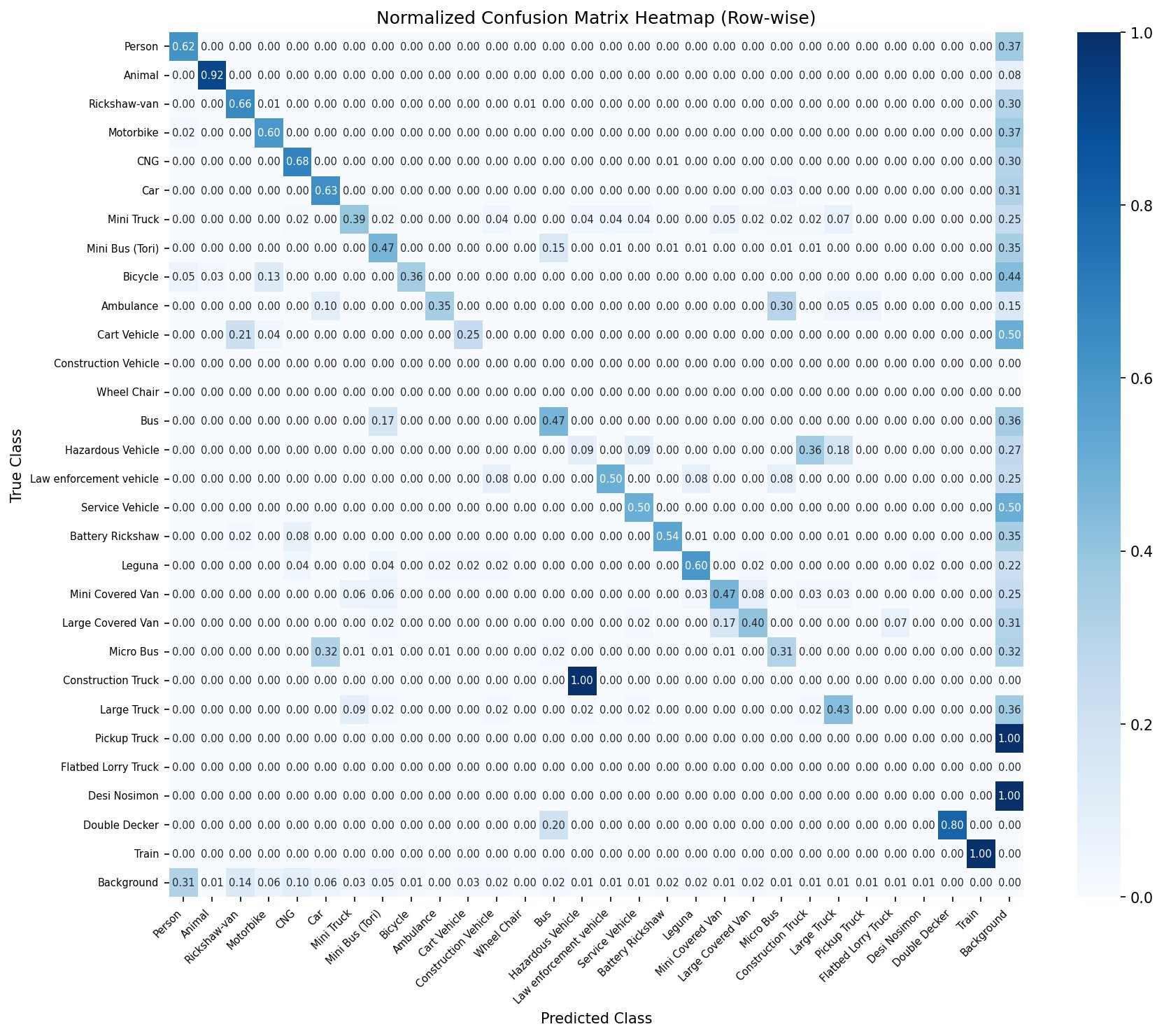}
        \caption{YOLOv8n}
        \label{fig:pred_v8n}
    \end{subfigure}
    \hfill
    \begin{subfigure}{0.3\linewidth}
        \includegraphics[width=\linewidth]{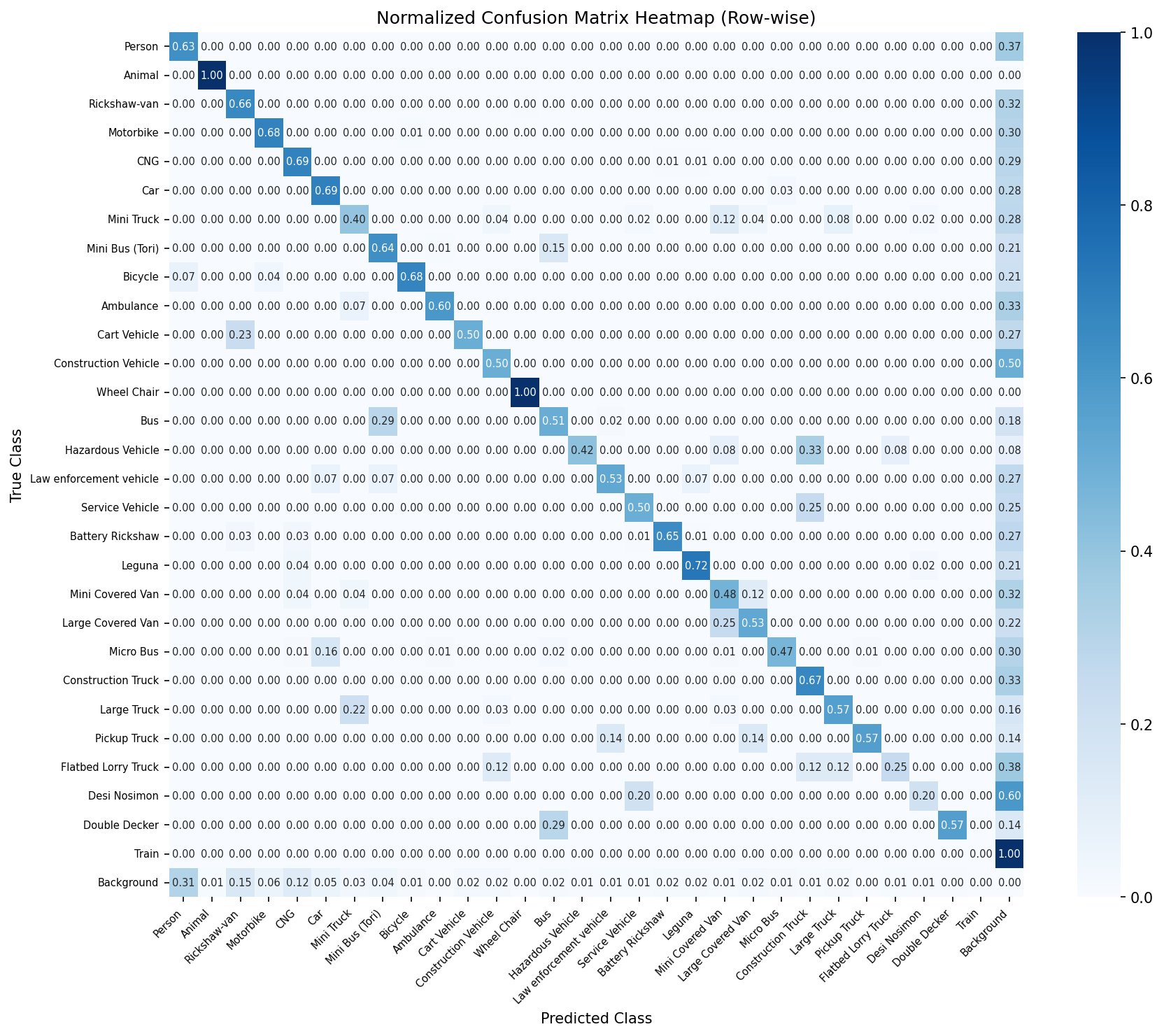}
        \caption{YOLOv8m}
        \label{fig:pred_v8m}
    \end{subfigure}
    \hfill
    \begin{subfigure}{0.3\linewidth}
        \includegraphics[width=\linewidth]{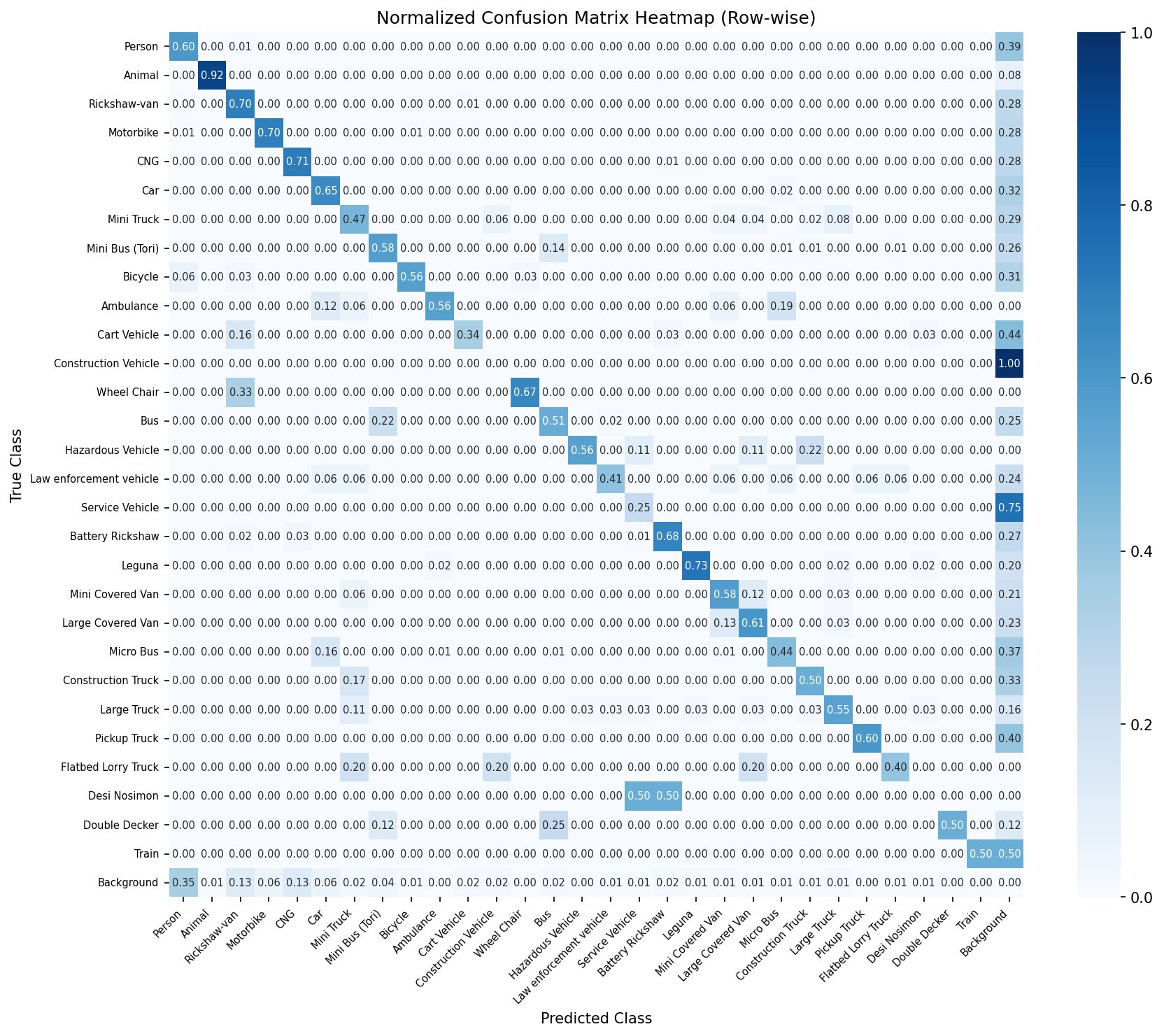}
        \caption{YOLOv8x}
        \label{fig:pred_v8x}
    \end{subfigure}
    \\[1em] 
    \begin{subfigure}{0.3\linewidth}
        \includegraphics[width=\linewidth]{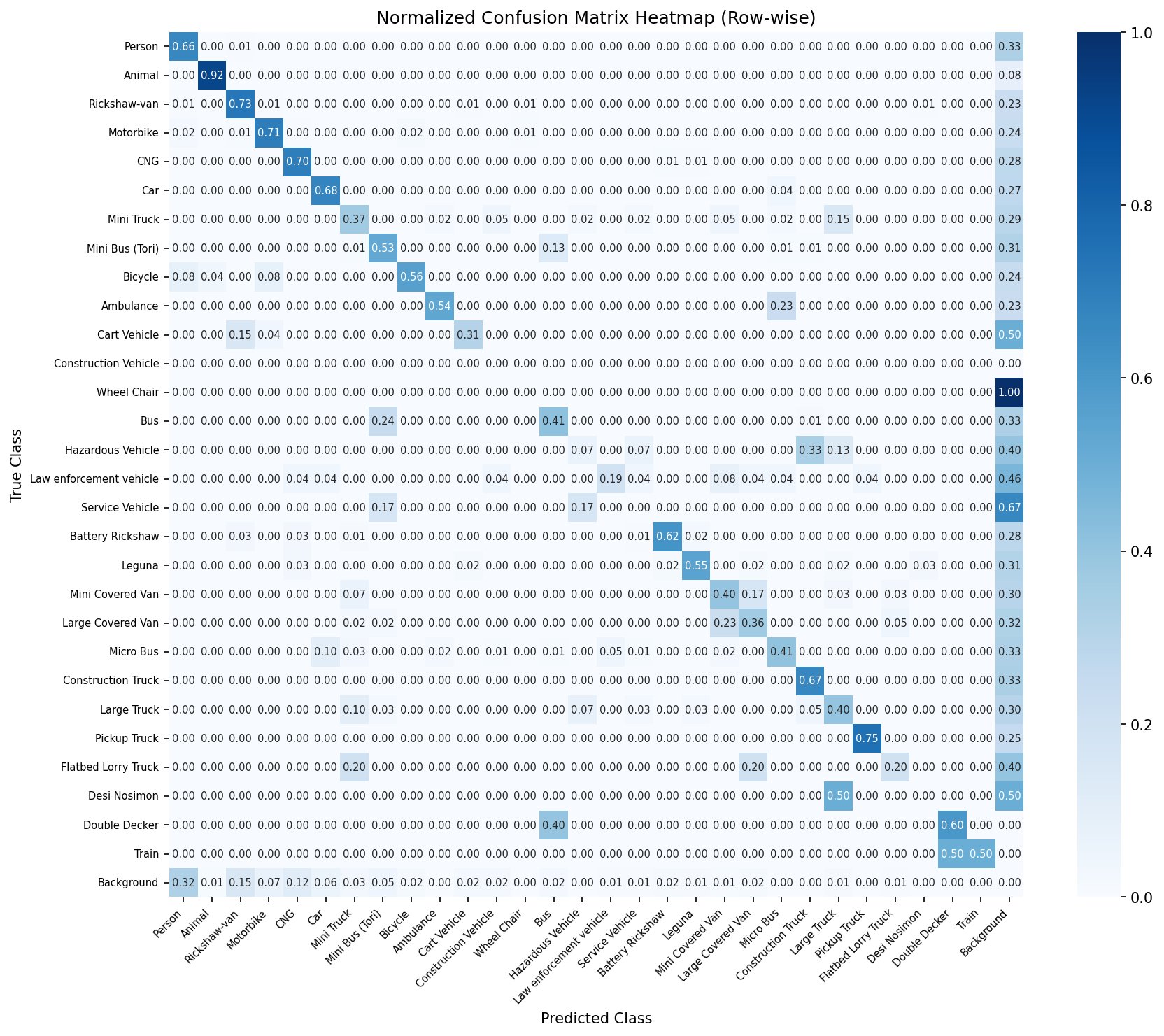}
        \caption{YOLOv11n}
        \label{fig:pred_v11n}
    \end{subfigure}
    \hfill
    \begin{subfigure}{0.3\linewidth}
        \includegraphics[width=\linewidth]{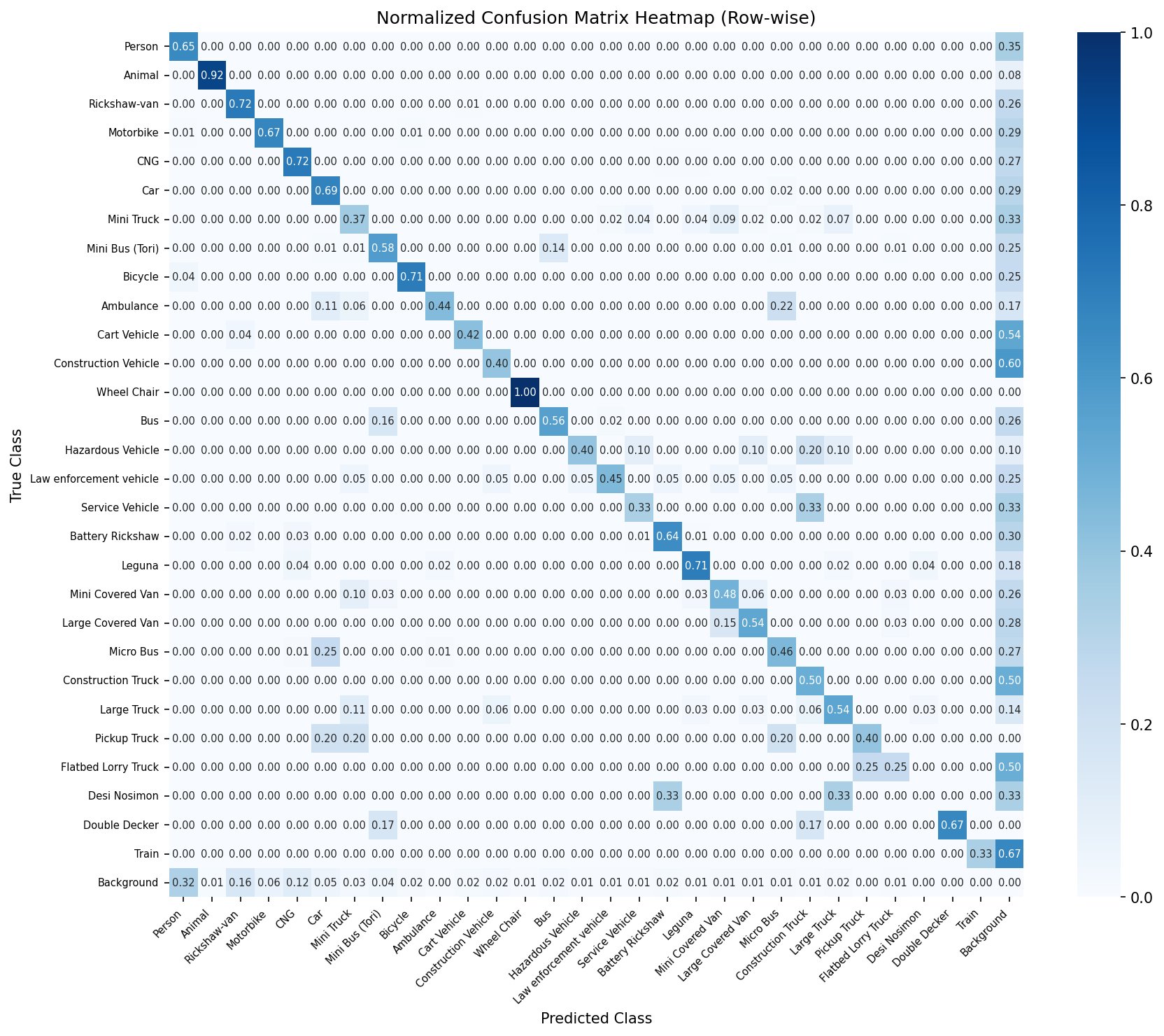}
        \caption{YOLOv11m}
        \label{fig:pred_v11m}
    \end{subfigure}
    \hfill
    \begin{subfigure}{0.3\linewidth}
        \includegraphics[width=\linewidth]{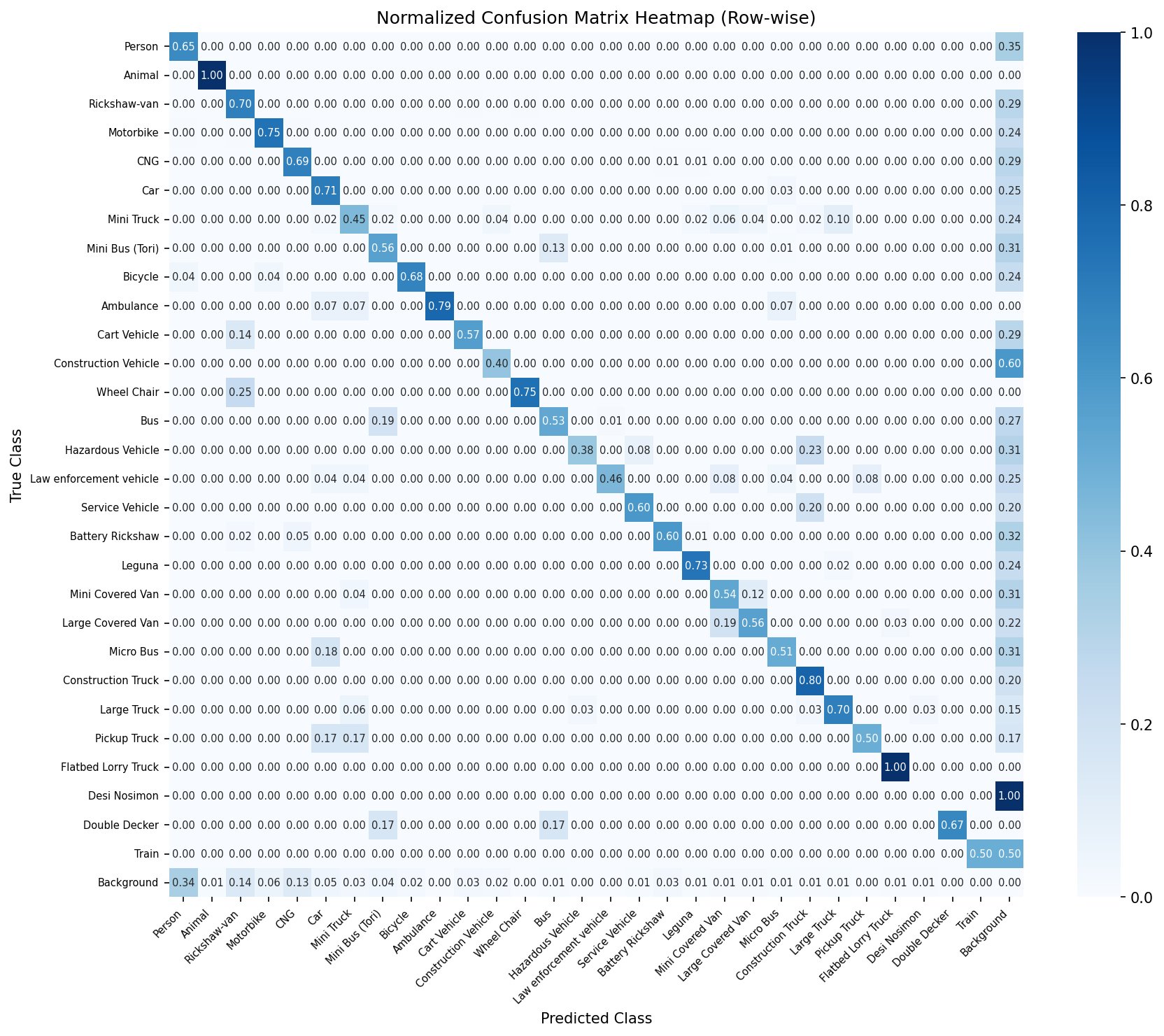}
        \caption{YOLOv11x}
        \label{fig:pred_v11x}
    \end{subfigure}
    \caption{Confusion Matrix For YOLO models used, showing Correct Prediction of Classes (Diagonal), And Missclassification(Off-diagonal)}
    \label{fig:yolo_model_confusion_matrix}
\end{figure}

\begin{figure}[htbp]
    \centering
    \begin{subfigure}{0.3\linewidth}
        \includegraphics[width=\linewidth]{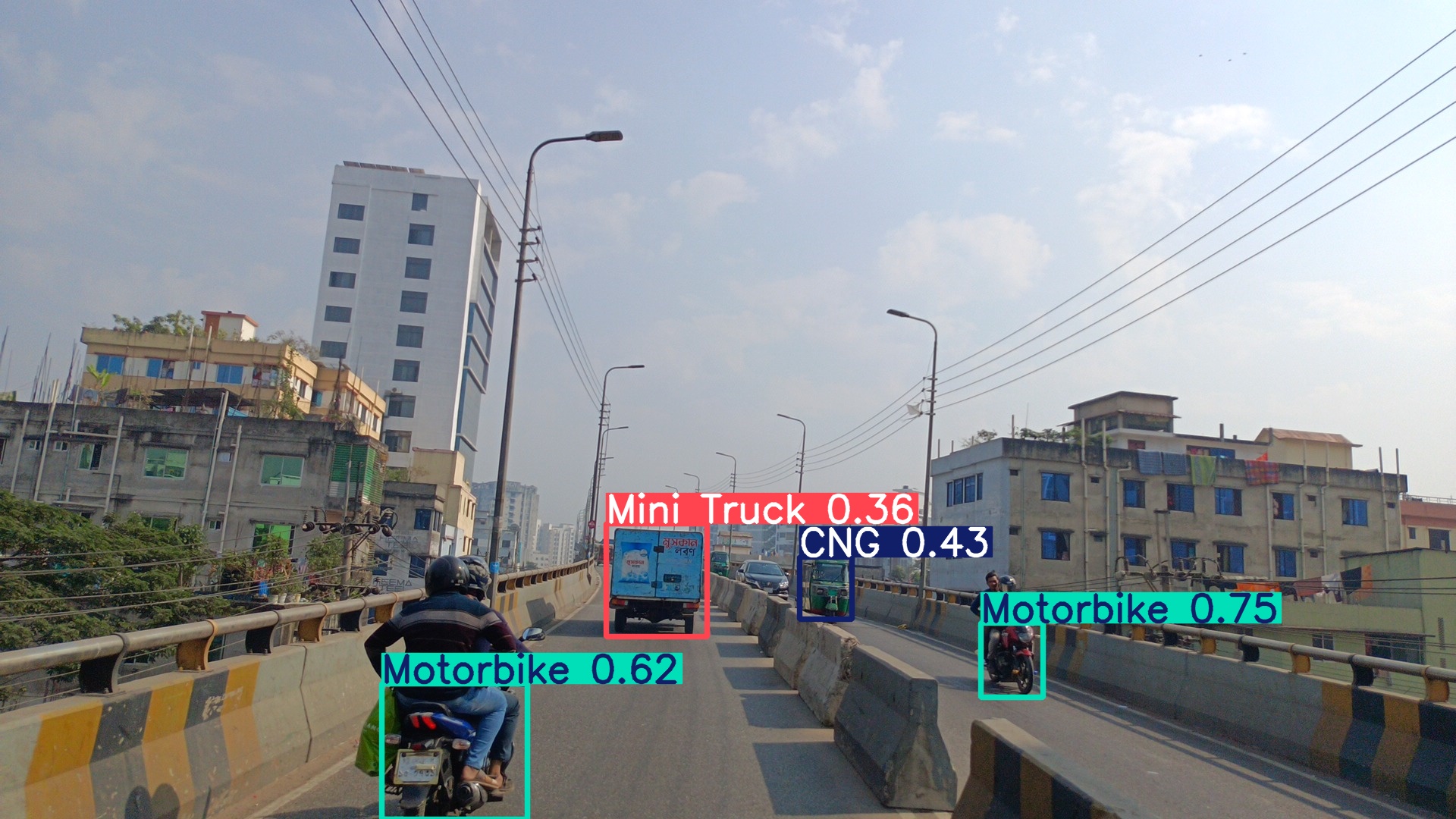}
        \caption{YOLOv8n}
        \label{fig:pred_v8n}
    \end{subfigure}
    \hfill
    \begin{subfigure}{0.3\linewidth}
        \includegraphics[width=\linewidth]{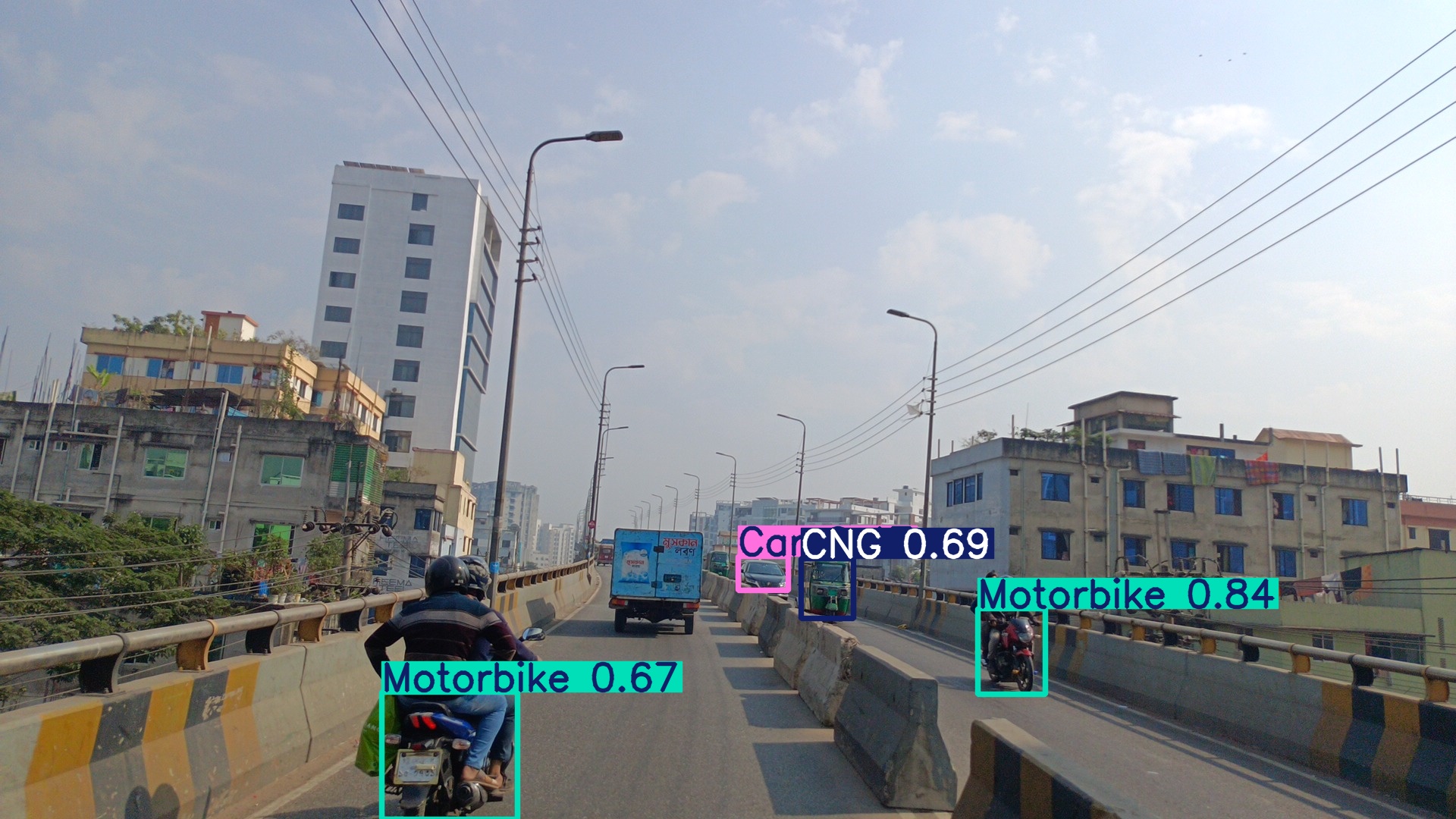}
        \caption{YOLOv8m}
        \label{fig:pred_v8m}
    \end{subfigure}
    \hfill
    \begin{subfigure}{0.3\linewidth}
        \includegraphics[width=\linewidth]{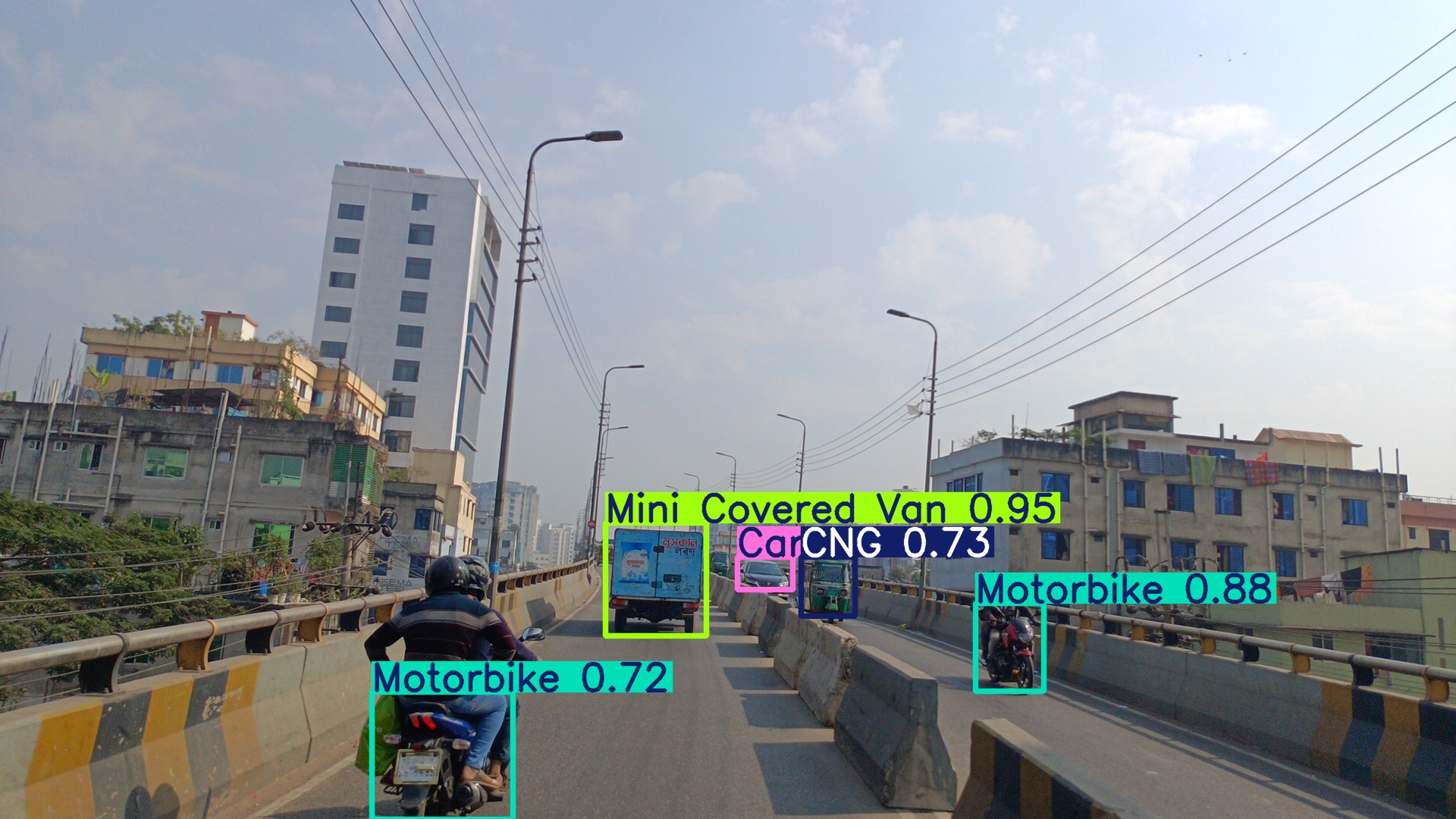}
        \caption{YOLOv8x}
        \label{fig:pred_v8x}
    \end{subfigure}
    \\[1em] 
    \begin{subfigure}{0.3\linewidth}
        \includegraphics[width=\linewidth]{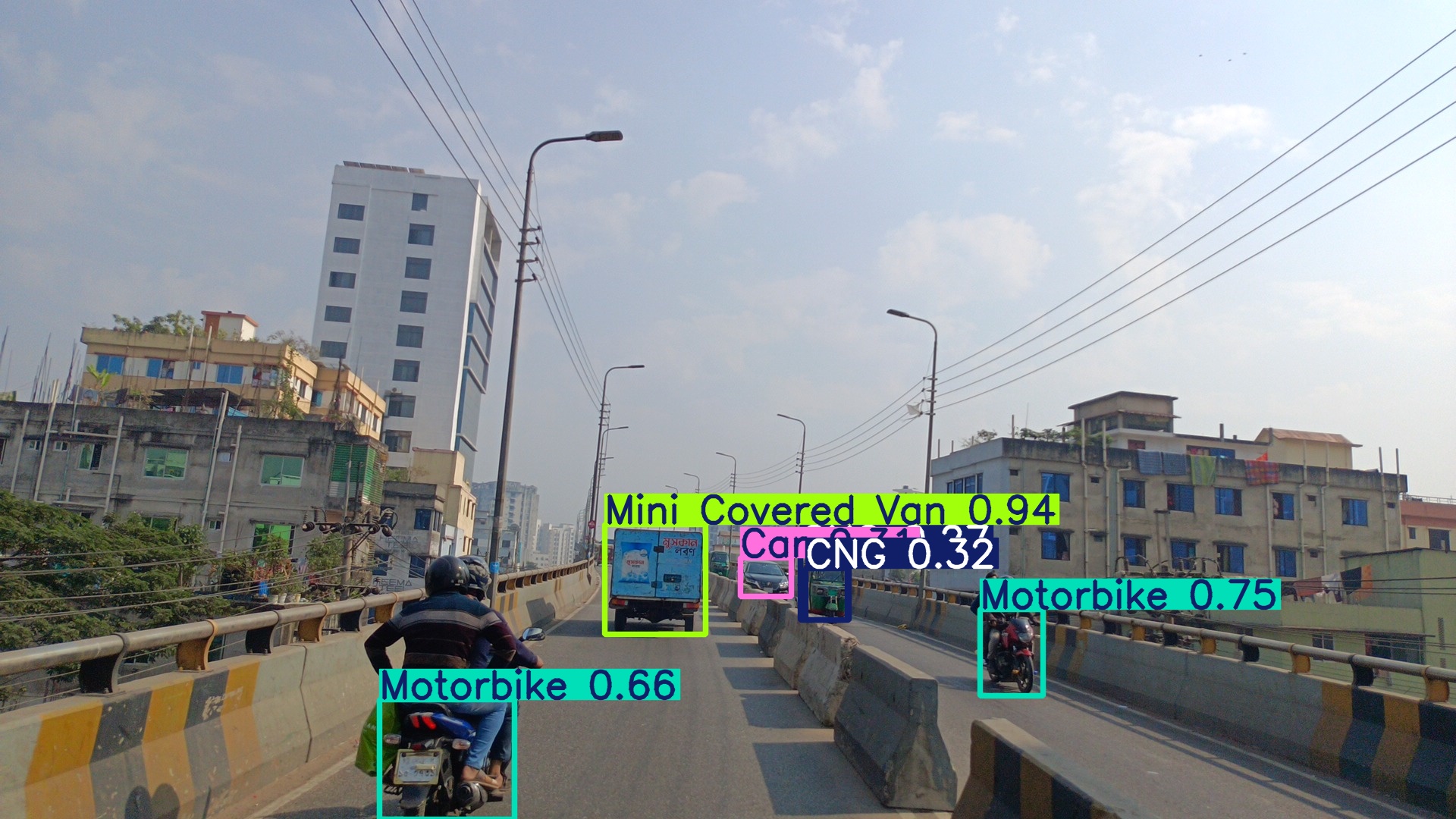}
        \caption{YOLOv11n}
        \label{fig:pred_v11n}
    \end{subfigure}
    \hfill
    \begin{subfigure}{0.3\linewidth}
        \includegraphics[width=\linewidth]{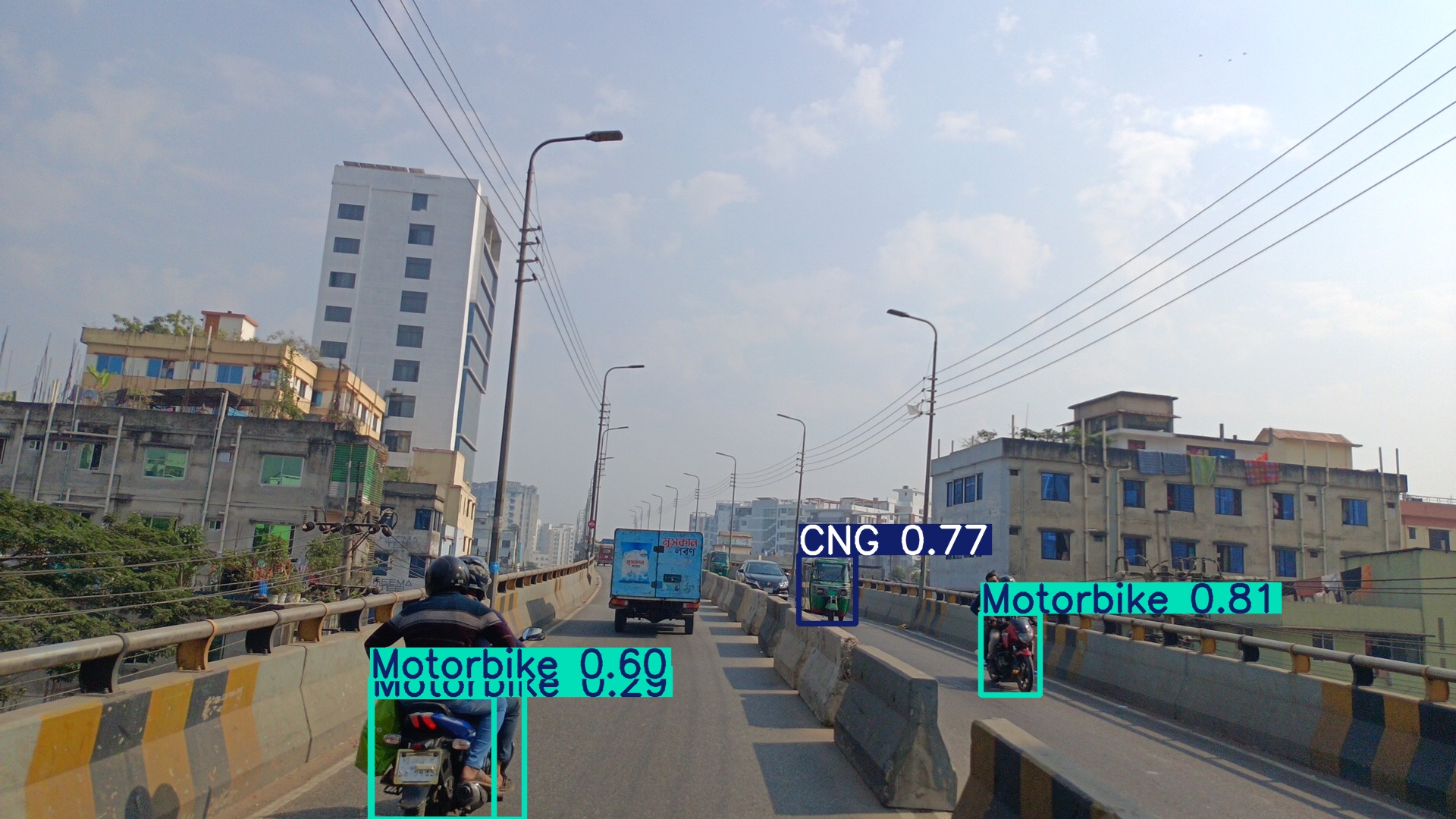}
        \caption{YOLOv11m}
        \label{fig:pred_v11m}
    \end{subfigure}
    \hfill
    \begin{subfigure}{0.3\linewidth}
        \includegraphics[width=\linewidth]{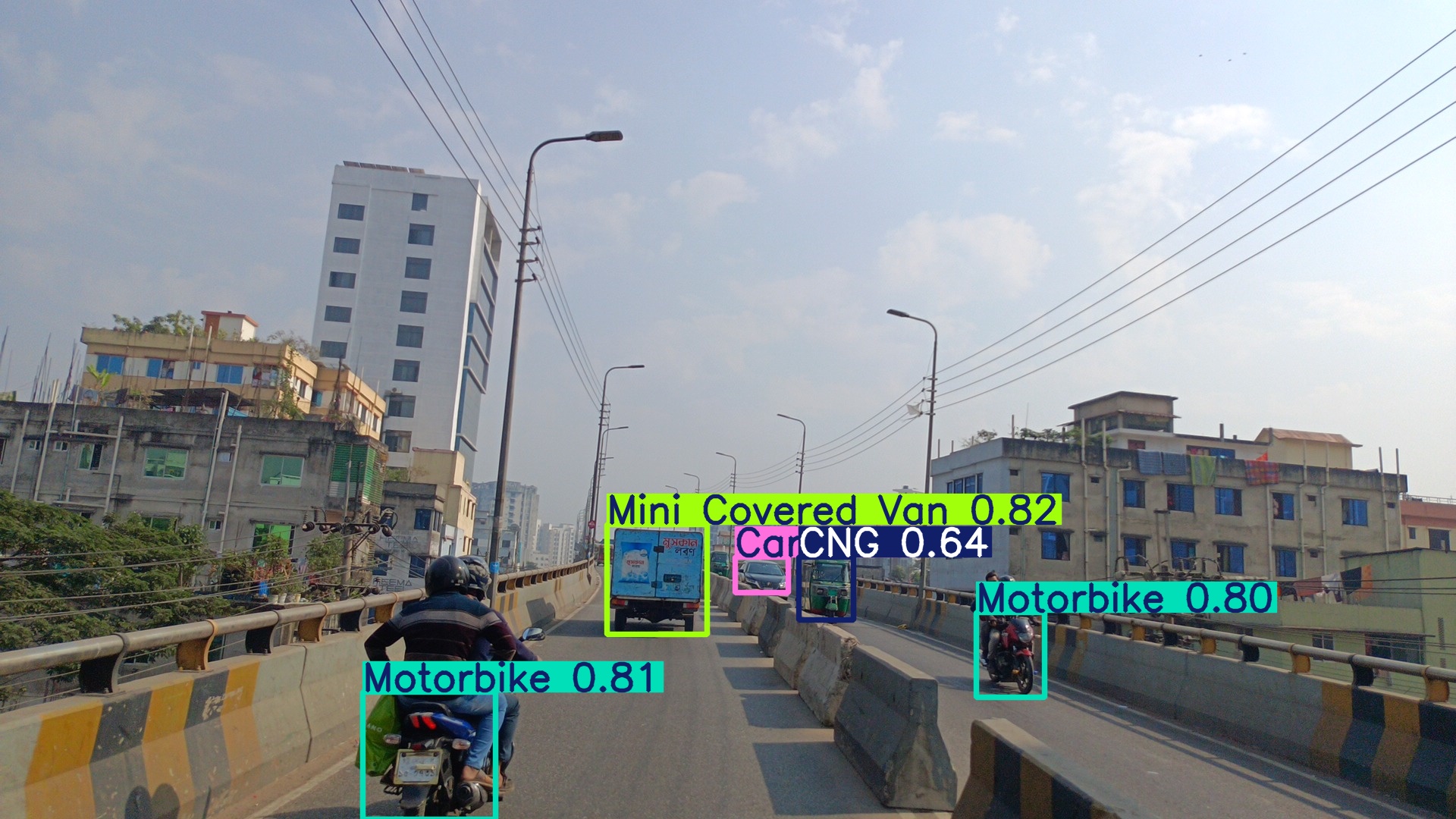}
        \caption{YOLOv11x}
        \label{fig:pred_v11x}
    \end{subfigure}
    \caption{Visualizations for six YOLO models, showing detected objects with bounding boxes, labels, and their confidence levels}
    \label{fig:yolo_model}
\end{figure}

Figure~\ref{fig:yolo_model_confusion_matrix} presents normalized confusion matrices which reveal critical insights into model performance. For common classes like Ambulance, both models achieve near-perfect precision (v8m: 89.5\%, v11x: 78.6\%) but show 0\% recall in confusion matrices, indicating rare detection despite high confidence when identified. This discrepancy arises because confusion matrices normalize by true instances---if a class is rarely detected (low recall), its diagonal value plummets even if predictions are accurate. Construction Vehicle and Desi Nosimon exhibit 0\% correct classification in both models, reflecting systemic failures for these rare classes. For visually similar pairs like Mini Truck and Mini Covered Van, confusion rates reach 30--40\% (v8m: 38.2\%, v11x: 34.7\%), highlighting persistent challenges in distinguishing shape-alike vehicles. The v11x model shows modest improvements: average correct classification rises to 59.2\% (vs. v8m's 54.4\%), and maximum confusion drops to 27.9\% (vs. v8m's 30.4\%). However, both models struggle equally with the rarest classes, where maximum misclassification rates hit 100\%---entire true classes being mislabeled. These metrics underscore the need for targeted data augmentation to address class imbalance and ambiguity.

\subsection{Model Performance}

Our analysis of all six model variants—detailed in Table~\ref{tab:model_metrics} and visualized in Figure~\ref{fig:yolo_comparison}—clearly demonstrates the trade-offs between detection accuracy and processing speed. The YOLOv11x (v11x) model leads with the highest mAP@0.5 (0.637), mAP@0.5:0.95 (0.438), recall (0.614), and F1 score (0.617), but requires over 45 milliseconds per image for inference. In contrast, the YOLOv8n (v8n) and YOLOv11n (v11n) models are much faster, processing images in just 2.5–2.7 milliseconds, but their detection metrics are noticeably lower (mAP@0.5 of 0.476–0.528). The YOLOv8m (v8m) and YOLOv11m (v11m) models strike a balance, offering robust detection performance (mAP@0.5 of 0.625–0.618) with moderate inference times of around 14–15 milliseconds. These results, summarized in Table 1 and highlighted in Figure 2, help guide model selection: choose v11x for maximum accuracy, v8n/v11n for real-time speed, and v8m/v11m for a practical middle ground. This balance is crucial for applications where both speed and precision matter.

\subsection{Critical Assessment}

Result analysis highlights key strengths and limitations of the evaluated models in detecting Bangladesh’s unique road objects. The YOLOv11 XL (v11x) model excels across most metrics, achieving the highest mAP@0.5 (63.7\%) and recall (61.4\%), particularly for common classes like Ambulance (AP@0.5=98.1\%) and Leguna (AP@0.5=84.5\%). Lightweight models (v8n, v11n) prioritize speed (~2.5 ms/image) but sacrifice accuracy, making them suitable for real-time applications despite lower detection rates. Mid-tier models (v8m, v11m) strike a practical balance, with mAP@0.5 values of ~62\% and moderate inference times (~15 ms/image).

However, significant challenges persist. Rare or visually ambiguous classes like Construction Vehicle (AP@0.5=0–29.8\%) and Desi Nosimon (AP@0.5=0–19.4\%) suffer from poor detection, reflecting dataset imbalances and insufficient training examples. Confusion matrices in figure~\ref{fig:yolo_model_confusion_matrix} reveal frequent misclassifications between similar vehicle types, such as Mini Truck vs. Large Covered Van and Service Vehicle vs. Pickup Truck, likely due to overlapping visual features in cluttered scenes. Additionally, the speed-accuracy trade-off—exemplified by v11x’s 45.8 ms/image vs. v11n’s 2.5 ms/image—limits real-world deployment for latency-sensitive applications like autonomous navigation. These limitations underscore the need for targeted improvements, particularly in addressing class imbalances and refining model architectures for Bangladesh’s dynamic roadscape.

%% file: sections/conclusion.tex
Bangladesh’s road environments are among the most unique and varied worldwide, which motivated this study to enhance and add value to vehicle detection research specifically tailored to these conditions. Our investigation revealed that YOLO models are most effective for real-time detection within such a diverse setting. Notably, YOLOv11-X achieved the highest accuracy despite a large class system where some classes had limited instances. However, the YOLOv8 and YOLOv11 models in their ``m'' configurations excelled in balancing accuracy and inference speed, both of which are essential for real-time vehicle detection. From this, we can conclude that among the 6 tested variants of ``YOLO'', the  ``m'' variants serve the best for this study's goal. This insight is significant, as real-time detection can be integrated into areas like autonomous driving and traffic law enforcement, directly impacting road safety—a key concern given the complexity of Bangladeshi roadscapes.

The future objective of this study is to further refine the dataset to enable models to interpret road scenes more effectively. Additionally, we aim to evaluate and compare other models with existing ones, striving to develop an optimal model paired with a robust dataset capable of accurately analyzing the entire Bangladeshi road environment. Such progress can support related research that urgently requires these advancements.